# An efficient hybrid classification approach for COVID-19 based on Harris Hawks Optimization and Salp Swarm Optimization

**Abubakr Issa,** University of Technology, Baghdad, iraq
**Yossra Ali,** University of Technology, Baghdad, Iraq
**Tarik Rashid,** University of Kurdistan Hewler, KRG, Iraq

**Abstract—** Feature selection can be defined as one of the pre-processing steps that decrease the dimensionality of a dataset by identifying the most significant attributes while also boosting the accuracy of classification. For solving feature selection problems, this study presents a hybrid binary version of the Harris Hawks Optimization algorithm (HHO) and Salp Swarm Optimization (SSA) (HHOSSA) for Covid-19 classification. The proposed (HHOSSA) presents a strategy for improving the basic HHO's performance using the Salp algorithm's power to select the best fitness values. The HHOSSA was tested against two well-known optimization algorithms, the Whale Optimization Algorithm (WOA) and the Grey wolf optimizer (GWO), utilizing a total of 800 chest X-ray images. A total of four performance metrics (Accuracy, Recall, Precision, F1) were employed in the studies using three classifiers (Support vector machines (SVMs), k-Nearest Neighbor (KNN), and Extreme Gradient Boosting (XGBoost)). The proposed algorithm (HHOSSA) achieved 96% accuracy with the SVM classifier, and 98% accuracy with two classifiers, XGboost and KNN.

**Keywords——** Feature selection, Hybrid Swarm intelligence, classification, Covid-19, medical image

## 1   Introduction

Medical image processing can be defined as one of the most significant areas in medical science, and it has a substantial effect on visualization applications. Also, medical image processing has a broad range of applications in medical diagnoses (treating and investigating diseases) and medical sciences (such as physiological and anatomical studies). Medical physics, medical engineering, biology, and optics are some of the fields of science that make up this medical science. With the discovery of X-rays, William Roentgen initiated the first efforts at contemporary medical imaging. Coronavirus (COVID-19), also known as SARS-Corona Virus-2, is a virus that results in causing severe acute respiratory syndrome (SARS-CoV2), is a viral infection that first occurred in Wuhan at the end of 2019. Due to such an outbreak, COVID-19 became a pandemic, threatening human lives and wreaking havoc on the economy. Therefore, many studies have been launched in an attempt to identify a way to restrict mortality and spread.

Those researches include the suggested treatment strategy, the screening method for early-stage patients, and the evaluation of different phases and recovery of treated patients. In hospitals, imaging techniques like chest X-rays are commonly utilized for detecting the severity and existence of COVID-19 pneumonia [1][2]. For improving the suggested system's training, X-ray images are often maintained in a medical database for subsequent investigation by multiple research organizations. Low contrast, noise, blurs, and faded colors are frequent problems, and images should be pre-processed to enhance quality by reducing noise.

The second stage is image segmentation, which depends on some attributes including color, texture, and depth measurements. The type of image and characteristics of the problem (disease) are chosen to determine which segmentation technique is used. The identification and extraction of features is the third stage. As the number of features that have been extracted from the image grows, the accuracy of classification decreases. In the classification vision, we can call it the curse of dimensionality. Feature optimization is a viable option for dealing with this issue.[3]

The 4th stage is the feature selection that has been obtained from the known properties using robust Optimization algorithms for better disease identifications from the medical images. The image was classified using one of the classifiers. Feature selection is a step in the preprocessing process that tries to increase the relevancy of obtained data by deleting irrelevant characteristics and choosing just relevant or useful variables [5]. Feature selection comprises reviewing feature subsets, employing certain search approaches to locate the best feature subset, assessing the chosen features, stopping criteria, and subset validation in general.[6]

There are three types of feature selection classifiers: wrapper schemes, filer schemes, and embedding schemes. The filter method, in contrast to the wrapper scheme, which is characterized by good classification accuracy and low speed, is rapid but inaccurate. The embedded system is preferred in the case when handling a certain model [7]. Filter techniques use the qualities of training data to assess the quality of features. Those approaches do not employ machine learning algorithms. Before choosing features with the highest score, filter methods usually take into account the score of all features. At the same time, other filtering approaches favor features with the greatest score per iteration [8]. Other well-known methods, like the correlation-based feature selection approach in [9] as well as dimensionality reduction methods and NNs in [10], can greatly decrease computational load and system complexity. Filter approaches overlook the performance regarding the chosen characteristics despite their speed and low computational cost [11].

Wrapper approaches utilize an evaluation algorithm to assess the specified features' quality. SVMs, Decision trees (DTs), KNN, Naïve Bayesian (NB), linear discriminant analysis (LDA), local neighborhood structure preserving embedding (LNSPE), artificial neural networks (ANNs), and local geometrical structure Fisher analysis (LGSFA) are some of the major wrapper's methods utilized for feature selection. In almost all cases, wrapper approaches outperform filter ones. Meta-heuristic algorithms are more advanced search algorithms that result from the evolution and expansion of feature selection. For instance, ongoing research to increase the performance regarding evolutionary algorithms (EA) like Genetic Algorithms (GAs) and Swarm Intelligence (SI) like Particle Swarm Optimization (PSO), Artificial Bee Colony (ABC), and Ant Colony Optimization (ACO) are underway. Grasshopper Optimization Algorithm (GOA), Grey Wolf Optimizer (GWO), Butterfly Optimization Algorithm (BOA), Harris Hawks Optimization (HHO) Whale Optimization Algorithm (WOA), and Ant Lion Optimization (ALO) are examples of recent algorithms. Metaheuristic algorithms are classified according to their exploration and exploitation phases into single solution based (i.e., Tabu Search (TS) and Simulated annealing (SA)) or population size based (in other words, GA, ACO, and PSO). The key contributions of this research are listed below:

• Suggest an effective hybrid classification method for COVID-19 with the use of the hybrid swarm algorithms (HHO, SSA).
This novel hybrid algorithm must improve resource consumption and performance, as well as storage capacity, reducing processing time.
• With the use of multiple classifiers (KNN, SVM, XGboost), test the suggested (HHOSSA) algorithm on datasets containing some positive negative COVID-19 chest X-ray scan images.
• Individual, hybridized predictor models and state-of-the-art techniques (WOA, GWO) are compared in terms of performance.

The sections of this paper are organized as follows: Section 2 provides a concise summary of some of the most related works. Section 3 discusses methodology. In section 4 we described in detail our proposed approach. Tools are illustrated in section 5. Performance evaluation is described in section 6. Results and discussion are included in section 7. Finally, the conclusions and future works are stated in section 8.

## 2    Related Works

Many studies have employed hybrid algorithms to handle a range of challenges recently. Hybrid algorithms have received a lot of attention lately, notably in feature selection optimization. Low-level hybrid algorithms and high-level hybrid algorithms are the 2 forms of hybrid algorithms. There are 2 types of hybridization schemes in high-level hybrid algorithms: high-level teamwork hybridization (HTH) and high-level rely-on hybridization (HRH). The self-contained meta-heuristics have been carried out in order in HRH, whereas in the HTH, one algorithm assists the other by supplying information via cooperative search. Low-level hybridization has been separated into two types: low-level teamwork hybrid (LTH) and low-level rely-on hybrid (LRH), both of which contain one meta-heuristic algorithm [12]. In the feature selection field, it has been observed that hybrid algorithms surpass native algorithms concerning performance. In the year 2004, the search process has been controlled by merging local search approaches with a GA algorithm, which was the first time a hybrid metaheuristics approach was utilized in feature selection. A combination with the EGA filter has been provided in a wrapper technique for text categorization [13]. A hybrid approach for feature selection has lately been created in various metaheuristic algorithms. In [13], the Binary Grey Wolf algorithm was combined with the Harris Hawks algorithm to create an excellent balance between exploitation and exploration to prevent local optimum solutions and increase solution precision. Harris Hawks was hybridized in [14] using Bitwise operations and Simulated Annealing for supporting the HHO algorithm's exploitation capacity and getting out of local optima. In [15], the Salp swarm algorithm was used to modify teaching–learning based optimization. This integration gives TLBO more flexibility in the exploration of population and achieving variety while also allowing it to swiftly attain the optimal value. They combined the Salp swarm algorithm with the Particle swarm algorithm in [16], in which the SSA was utilized for updating the salps positions and the PSO was utilized otherwise. This hybridization was utilized for the improvement of the exploration and exploitation of the Salp swarm algorithm.

## 3       Methodology

### 3.1 Harris Hawks optimization algorithm

HHO can be defined as one of the swarm metaheuristic algorithms inspired by Harris Hawks' hunting behavior of "seven kills" or "surprise pounce." Based on the prey's fleeing behavior nature, hunting duration can range from some seconds to many hours. The modeling algorithm of HHO is split into 2 parts (exploitation and exploration). Harris' hawks have been employed as candidate solutions in the HHO algorithm, with the best candidate solution reflecting the desired or optimum prey in each stage [17]. The first phase pertains to the process of perching and detection of the prey. The algorithm simulates Harris' hawks' perching methods in 2 separate scenarios. Harris' hawks are assumed to perch in various locations during their group home range in the first scenario. In Eq (1), q=0.50 models that condition.

$$\overrightarrow{X_1}(t+1)=\begin{cases} X_{rand}(t) - r1|X_{rand}(t) - 2r_2X(t)|, q \geq 0.50 \\ (X_{rabbit}(t) - X_m(t)) - r_3(LB + r_4(UB - LB)), q < 0.50 \end{cases} \quad (1)$$

While the other likelihood is that Harris' hawks would perch on positions near other swarm members and prey. This condition has been introduced in Eq1 for q < 0.50:

where $\vec{X_1}(t+1)$ is Hawks' position vector, t represents the following iteration, $X_{rand}(t)$ is a hawk that has been chosen at random from the current population, $X(t)$ represents the position vector of hawks, r1, r2, r3, r4, and q represent random numbers in the range of (0,1), $X_{rabbit}(t)$ represents rabbit position, $X_m$ denotes the average position of the current population of the hawks, lower and upper bounds for generating random locations inside the Hawks' stadium are Lb and Ub, respectively.

While in the phase of exploitation, the Harris' hawks attack prey which has been identified in the preceding step. The algorithm has 4 different possibilities for modeling various attacking styles that have been utilized by Harris' hawks.

While r denotes the probability of prey escaping, successful escape has been donated by r < 0.50, whereas r ≥ 0.50 denotes failure to escape. Depending upon the prey's chances of escaping (r), hawks will use either soft or hard besieges to catch prey. The algorithm's parameter E has been utilized for the determination of the type of attacking besieges. If the prey is unable to escape when r ≥ 0.50 hard besiege happens when |E| < 0.50 and soft besiege takes the place in the case where |E|≥ 0.50 The mathematical Modelling of soft besiege has been represented by Eqs (2) through (3), and hard besiege has been shown by Eq (4):

$$X(t+1) = \Delta X(t) - E|J x X_{rabbit}(t) - X(t)| \quad (2)$$
$$\Delta(t) = X_{rabbit}(t) - X(t) \quad (3)$$
$$X(t+1) = X_{rabbit}(t) - E|\Delta X(t)| \quad (4)$$

In the case of successful escaping of the prey (r<0.50), soft besiege with a progressive rapid dive take is applied in the case where |E|≥ 0.50 as shown in Eq (5), Eq (7), Eq(8) while Hard besiege with the progressive fast dive occurs in a case where |E|≥ 0.50 as shown in Eqs (6), (7), and (8):

$$Y = X_{rabbit}(t) - E|J * X_{rabbit}(t) - X(t)| \quad (5)$$
$$Y = X_{rabbit}(t) - E|J * X_{rabbit}(t) - X_m(t)|, X_m(t) = \frac{1}{N} * \sum_{i=1}^{N} X_i(t) \quad (6)$$
$$Z = Y + S \times LF(D) \quad (7)$$
$$X(t+1) = \begin{cases} Y, if\ f(Y) < F(X(t)) \\ Z, if\ f(Z) < F(X(t)) \end{cases} \quad (8)$$

D represents the problem dimension and S represents the random vector by 1xD size and LF represents the function of levy flight, estimated with the use of Eq. (9):

$$LF(x) = 0.01 \times \frac{u \times \sigma}{|v|^{\frac{1}{\beta}}}, \sigma = \left(\frac{\Gamma(1+\beta) \times \sin(\pi\beta/2)}{\Gamma\left(\frac{1+\beta}{2}\right) \times \beta \times 2^{\left(\frac{\beta-1}{2}\right)}}\right) \quad (9)$$

**The energy of a rabbit is modeled as $E = 2E_0 \left(1 - \frac{t}{T}\right)$ ( 10 )**

Where E represents the prey's escaping energy, T represents the maximal number of iterations, and $E_o$ represents its initial energy state.

### 3.2 Salp swarm optimization algorithm

SSA can be defined as a swarm metaheuristic algorithm [18] that was created for solving various optimization problems. It was inspired by the activity of Salps in nature; salps are a type of jellyfish with tissues comparable to jellyfish and a high water percentage in their moving behavior and weights [19]. They move by contracting their bodies and shifting positions by pumping water through them. The salp chain describes the swarming behavior of salps in the ocean. By allowing for faster and more harmonic changes, this behavior could benefit salps in foraging and better movement. [18] Salp chains were theoretically modeled and after that tested in optimization problems as a result of this characteristic [16]. The algorithm starts its work by dividing the generated population into 2 parts (which are: leader and followers) where the leader leads the salp chain and the remaining salps play the role of followers. A salp uses the food source as a target in an n-dimensional search space. The following equation has been used to update the leader's position:

$$X_j^1 = \begin{cases} F_j + r_1\left((Vmax_j - Vmin_j)r_2 + Vmin_i\right), r_3 \geq 0 \\ F_j - r_1\left((Vmax_j - Vmin_j)r_2 + Vmin_i\right), r_3 < 0 \end{cases} \quad (11)$$

Where $X_j^1$ represent the position of leader in the j[th] dimension and $F_j$ is food's location. The upper is represented by $Vmax_j$ and the lower bounds that have been denoted by $Vmin_j$. The search space is maintained using the 2 random variables $r_2$ & $r_3$ in the range [0, 1].

The parameter $r_1$ is also an important control parameter in the process of exploration and exploitation and it is calculated by using Eq (12).

$$r_1 = 2e^{\left(\frac{-4t}{N}\right)^2} \quad (12)$$

Where t represents the current iteration and N denotes the maximum amount of iterations. In a case where the position of the leader has been changed, Eq (13) is used to change the followers' position:

$$X_j^i = \frac{1}{2}(X_j^1 - X_j^{i-1}) \quad (13)$$

**Where $X_j^i$ denotes the ith follower's position in the jth dimension, and the value of I must be > 1.**

## 4 The proposed approach

Despite its simple structure and fast convergence rate, the HHO algorithm is not without flaws. However, in the domain of feature selection optimization, the algorithm may encounter a balancing problem between the exploration and exploitation phases, resulting in a local optimum. Problems can arise during the feature selection process when dealing with the high-dimensional feature set. In general, the HHO algorithm optimization power depends on the best optimal solution selected based on the best fitness value. In this paper, we present a strategy for improving the basic HHO's performance using the Salp algorithm's power to select the best solution.

### 4.1 The structure of HHOSSA

The proposed hybrid algorithm HHOSSA contains many stages: Initialization and binarization function, Best fitness selection, and Evaluation.

### 4.2 Initialization and binarization function

In this phase, the HHO algorithm generates a random initial population X that contains k Hawks which is every k represents a new solution this vector of d dimension of features and using binary representations of (0 and 1) to represent the selected features where every feature that selected will represent by 1 and every refused feature will represent by 0 by using of the following binarization function:

$$X_{binary} = \begin{cases} 1 \text{ if } x > thre\_val \\ 0 \text{ if } x < thre\_val \end{cases} \text{ where } thre\_val = 0.5 \quad (14)$$

### 4.3 Best fitness selection

In basic HHO the position vectors Xrand and Xrabbit are responsible for the exploration step that has been characterized by Eq1, which is critical for balancing the exploitation and exploration phases. Position vectors with higher significance speed up global exploration, while those with lower significance speed up exploitation. As a result, an appropriate Xrand and Xrabbit selection should be made to achieve a stable balance between local exploitation and global exploration [20]. In this phase, the SSA algorithm will be used to find a better solution where the SSA algorithm finds the new fitness and if the new one is better than the one that has been found by the HHO algorithm so the new one will be replaced and the Xrabbit will be changed also otherwise, the HHO solution remains unchanged.

The goal of feature selection is to reduce the number of features and classification error rate, i.e., through the removal of the redundant and irrelevant features and keeping the relevant ones only, classification accuracy is improved. The KNN classifier was used in this study because it is simple to evaluate the fitness function Eq (15), which was used, expresses the fitness function that was used.

$$Fitness = a * class_{err} + b * (\frac{f_{sel}}{f_{max}}) \quad (15)$$

Where a =0.9 is constant for controlling the accuracy, b=[0.1, a] random number enhances the accuracy, $class_{err}$ is the rate of classification error and $f_{sel}$ represents the number of the selected feature and $f_{max}$ represents the total amount of features.

**Algorithm1 Pseudo-Code of HHOSSA Algorithm**

Input: H population size, T iteration number, ub=1, lb=0, thre_val=0.5, levy_beta=1.5
Output: Best selected features vector

Randomly initialize of population H random hawks $x_i$ (i=1,2,3,….., H)
Compute the fitness value of every one of the hawks **$F_{hho}$**
$X_{rabbit}$ = best solution found

**While** (the stop condition isn't met) **do**
Compute the fitness values of the hawks
Set $X_{rabbit}$ as rabbit location (i.e. optimal location)
**For** (each hawk ($X_i$)) **do**
 Update ($E_o$ , J)
 **if** (|E| ≥ 1) **then**
 Update location vector according to Eq1
 **if** (|E| < 1) **then**
 **if** (r ≥0.50 & |E| ≥ 0.50 ) **then**
 Update location vector through utilizing Eq. (2)
 **else if** (r ≥0.50 & |E| < 0.50 ) **then**
 Update location vector through utilizing Eq. (4)
 **else if** (r <0.50 & |E| < 0.50 ) **then**
 Update location vector through utilizing Eq. (8)
 **else if** (r <0.50 & |E| < 0.50 ) **then**
 Update location vector through utilizing Eq. (8)
Apply the SSA algorithm to find the best fitness **$F_{ssa}$** using Eq. (15)
If (**$F_{ssa}$** < **$F_{hho}$** )
Update ($X_{rabbit}$, $X_{rand}$ )
**End if**
    **End While**

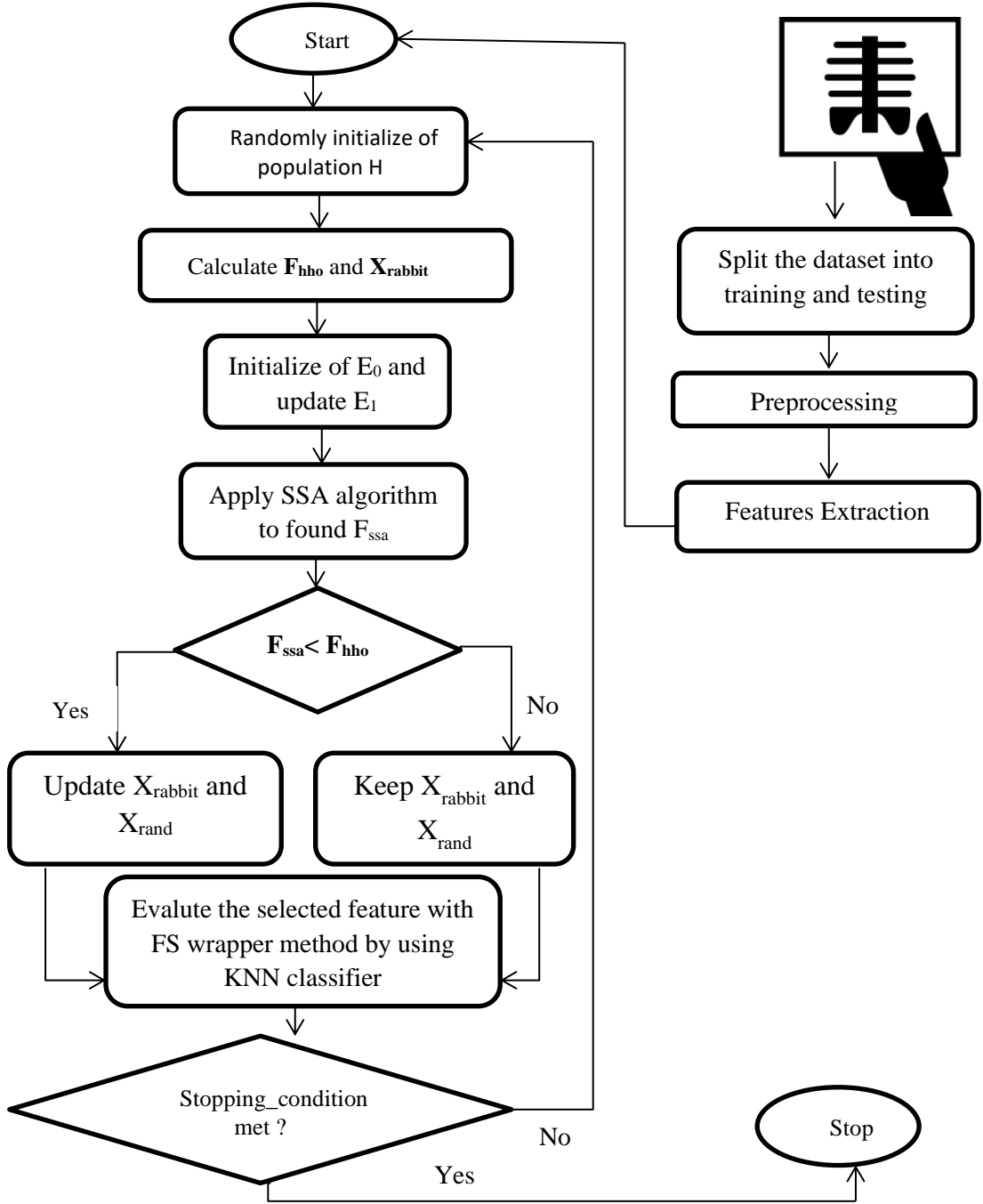

Figure 1: Structure of the proposed HHOSSA algorithm

## 5    Tools

### 5.1 Dataset

We are working with a dataset of 800 chest X-ray images obtained from [21-25]. This dataset consists of 400 chest X-ray images with confirmed COVID-19 infection, and 400 chest X-ray images of normal condition. This dataset images come with PNG file format and grey level scale and all images are resized to 200 × 200 pixels.

### 5.2 Classifiers

The main goal of classification is to categorize new samples that haven't been labeled for a particular class. However, we must first train the classifier for it to recognize the characteristics of the data, as well as the relationship between attribute values and the class label. Three classifiers are used in the methodology presented in this paper. The first one K-nearest neighbor classifier and it's used for the reasons of its straightforward implementation, with only one parameter K denoting the number of neighbors, which makes it more useful for identifying the best subset of attributes [26]. The second one is the SVM classifier which is a well-known constructive learning technique that is formalized by a separating hyperplane. Making a nonlinear transformation of the original input set to the high-dimensional set of features, where the optimum separating hyperplane may be found, can lead to a solution [27]. The third classifier is Extreme Gradient Boosting (XGBoost) which is a machine learning method that has been used for solving supervised learning problems. It has excellent scalability and a fast running speed, making it a popular machine-learning method [28].

## 6    Performance evaluation

The metrics of evaluation that are used to measure classification performance in this study are accuracy, precision, recall, and F1 as defined below:

$$Accuracy = \frac{TP+TN}{TP+TN+FP+FN} \; (16)$$

$$precision = \frac{TP}{TP+FP} \; (17)$$

$$recall = \frac{TP}{TP+FN} \; (18)$$

$$F1 = 2 \times \frac{Specificity \times Recall}{Specificity + Recall} \; (19), \; specificity = \frac{TN}{TN+FP} \; (20)$$

In which "TP" (true positives) denotes positive COVID-19 images which the classifier accurately labeled, and "TN" (i.e. true negatives) corresponds to the negatives COVID-19 images that have been successfully labeled by the classifier. False positives (FP) are positive COVID-19 images mislabeled as negative, whereas false negatives (FN) are negative COVID-19 images that have been incorrectly identified as positive COVID-19 images [29].

# 7 Results and discussion

A total of 800 X-ray images (400 covid-19 and 400 normal) have been collected from the digital database and utilized for testing the efficacy of the suggested hybrid approach, which utilized two state-of-art algorithms (SSA, HHO) for feature selection to improve the classification of the covid-19 infection with the use of automatic AI techniques and showed a high level of classification accuracy following testing and training. The dataset was divided into two sections: 20% for validation and testing and 80% for training. Table 2 demonstrates that the suggested hybrid method has a high accuracy percentage based on the classifiers utilized. The parameter setting for the suggested methodology has been listed in Table 1.

Table1: Parameter values for used methods

| **Methods** | **Parameter values** |
|---|---|
| HHOSSA algorithm | Feature size: 126<br>Population size: 30<br>Number of iterations for HHO:100<br>Number of iterations for SSA:20<br>Ub:1<br>Lb:0<br>Thre_val:0.5<br>Beta:1.5<br>Random variables a and b: 0.9, [0.1, a ] |
| KNN classifier | K=5<br>Classes count:2<br>No.of training set:224 |
| SVM classifier | Classes count:2<br>No.of training set:224 |
| XGboost classifier | Classes count:2<br>No.of training set:224 |

Table 2: Performance of HHOSSA over three classifiers KNN, SVM, and XGboost.

| **Classifier** | **Accuracy** | **Precision** | **Recall** | **F1** |
|---|---|---|---|---|
| KNN | 98.21428571428571 | 0.97 | 0.99 | 0.98 |
| SVM | 96.42857142857143 | 0.96 | 0.96 | 0.96 |
| XGboost | 98.21428571428571 | 0.99 | 0.96 | 0.98 |

### 7.1 Comparative study

The suggested system's performance was assessed utilizing a variety of modern optimization methods (GWO, WOA). Table (3) shows the performance of the HHO algorithm used for feature selection and gets 94%,89%, and 94% over three classifiers KNN, SVM, and XGboost, while Table(4) shows the performance of the SSA algorithm used for feature selection and gets 96%,80%,94% over three classifiers KNN, SVM, XGboost, Table (5) shows the performance of GWO algorithm used for feature selection and gets 96%,82%,92% over three classifiers KNN, SVM, XGboost, While Table (6) shows the performance of WOA algorithm used for feature selection and gets 96%,86%,96% over three classifiers KNN, SVM, XGboost.

Table 3: Performance of HHO over three classifiers KNN, SVM, and XGboost.

| Classifier | Accuracy | Precision | Recall | F1 |
|---|---|---|---|---|
| KNN | 94.64285714285714 | 0.90 | 0.99 | 0.95 |
| SVM | 89.28571428571429 | 0.87 | 0.93 | 0.90 |
| XGboost | 94.64285714285714 | 0.93 | 0.96 | 0.95 |

Table 4: Performance of SSA over three classifiers KNN, SVM, and XGboost.

| Classifier | Accuracy | Precision | Recall | F1 |
|---|---|---|---|---|
| KNN | 96.64285714285714 | 0.93 | 0.96 | 0.95 |
| SVM | 80.35714285714286 | 0.81 | 0.79 | 0.80 |
| XGboost | 94.64285714285714 | 0.96 | 0.93 | 0.95 |

Table 5: Performance of GWO over three classifiers KNN, SVM, and XGboost.

| Classifier | Accuracy | Precision | Recall | F1 |
|---|---|---|---|---|
| KNN | 96.42857142857143 | 0.93 | 0.99 | 0.97 |
| SVM | 82.14285714285714 | 0.74 | 0.99 | 0.85 |
| XGboost | 92.85714285714286 | 0.90 | 0.96 | 0.93 |

Table 6: Performance of WOA over three classifier KNN, SVM, and XGboost.

| Classifier | Accuracy | Precision | Recall | F1 |
|---|---|---|---|---|
| KNN | 94.64285714285714 | 0.90 | 0.99 | 0.95 |
| SVM | 89.28571428571429 | 0.87 | 0.93 | 0.90 |
| XGboost | 96.42857142857143 | 0.99 | 0.93 | 0.96 |

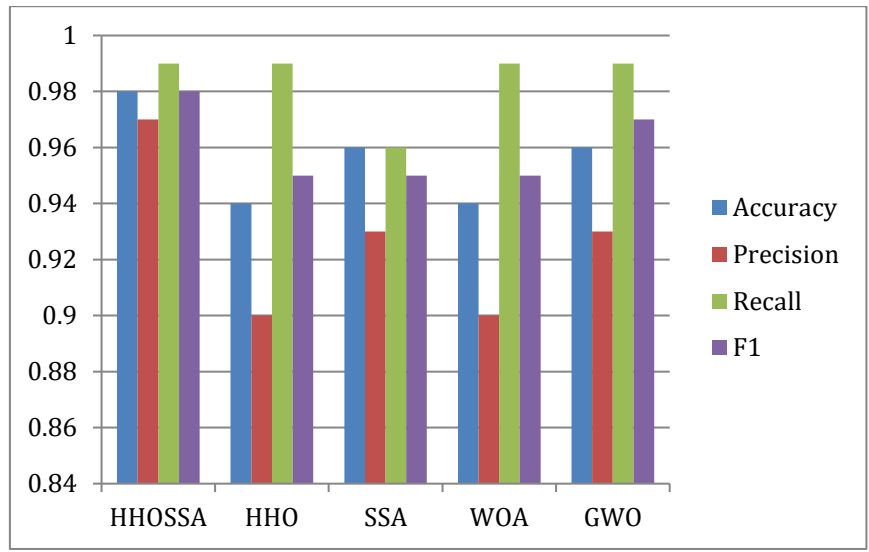

Figure 2: The accuracy, precision, recall, and the F1 values for all algorithms over the KNN classifier

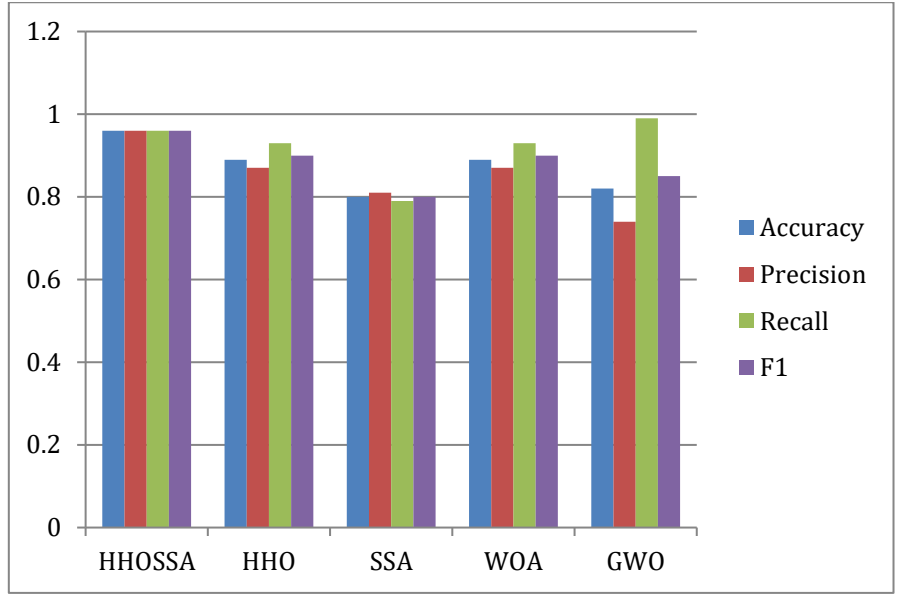

Figure 3: The accuracy, precision, recall, and the F1 values for all algorithms over the SVM classifier

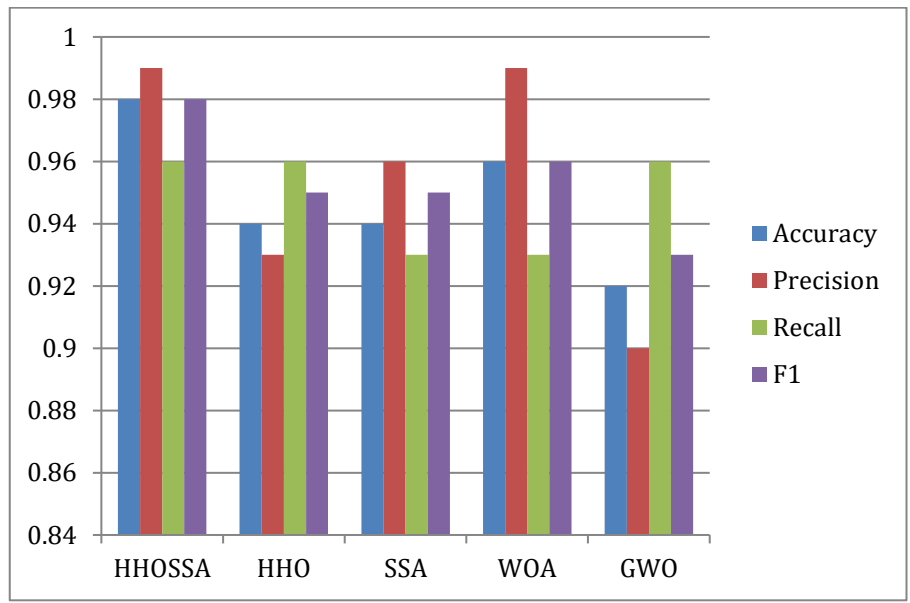

Figure 4: The accuracy, precision, recall, and the F1 values for all algorithms over the XGboost classifier

### 7.2 Software and Hardware Requirements

The proposed system operates by using a personal computer Lenovo that has specifications such as Intel(R) Intel(R) Core(TM) i7-6500U @ 2.59 GHz for CPU, 8 GB windows10 of RAM, and 64-bit Operating System. The proposed system is operated by using python 10 languages with (Pycharm) IDE. Table (7) shows the processing time of the proposed algorithm and stand-alone algorithms depending on the classification processing time of the testing dataset.

Table 7: Processing time of proposed (HHOSSA), HHO, SSA.

| Algorithm | Total processing time (seconds) |
|---|---|
| HHOSSA | 1.0661 |
| HHO | 0.9906 |
| SSA | 1.1425 |

It should be noted that the hybrid algorithm's processing time for completing the classification process is less than the sum of the processing times for the Harris hawk and Salp algorithms because the Salp algorithm's iterations are fewer than those of the Harris hawk algorithm within the hybrid algorithm. However, this improved the classification process and accelerated performance without degrading the hybrid algorithm's quality.

## 8   Conclusion and future works

The presented work presents a new hybrid swarm algorithm (referred to as HHOSSA) that combines the SSA and HHO for selecting the best features subset to improve the detection and classification of the COVID-19 virus with the use of chest X-ray images. The novel method provided to improve the process of the feature section and also for achieving the balance between exploitation and exploration of the HHO algorithm with the use of the capability of SSA for finding the best features subset It is noted that the processing time required to complete the classification process using the hybrid algorithm is less than the sum of the processing time of the Harris hawk and Salp algorithms because the number of iterations of the Salp algorithm is less than the iterations of Harris hawk algorithm inside hybrid algorithm, However, this did not affect the quality of the hybrid algorithm, but rather it increased the speed of performance and improved the classification process. A total of 800 (400 covid-19 and 400 normal) X-ray images are taken from the digital database to assess the HHOSSA's performance. XGboost and KNN classifiers get 98% accuracy, whereas SVM classifiers score 96%. We want to adapt the suggested technique to more applications in the future, including signal processing and cloud computing task scheduling. Furthermore, the HHO algorithm's searching power was used to construct a novel suggested algorithm in several aspects.

## 9       Acknowledgment

The authors would like to thank the University of Technology, Baghdad, Iraq for their continuous support for this research work.

## 10      References

[1] Lakshmanarao, A., Raja Babu, M., & Srinivasa Ravi Kiran, T. (2021). An Efficient Covid19 Epidemic Analysis and Prediction Model Using Machine Learning Algorithms. International Journal of Online and Biomedical Engineering (iJOE), 17(11), pp. 176–184. https://doi.org/10.3991/ijoe.v17i11.25209

[2] S. Cho, S. Lim, C. Kim, S.Wi, T. Kwon,W. S. Youn, S. H. Lee, B. S. Kang, and S. Cho, ``Enhancement of soft-tissue contrast in cone-beam CT using an anti-scatter grid with a sparse sampling approach,'' Phys. Medica, vol. 70, pp. 1_9, Feb. 2020

[3] Allam, Mohan, and M. Nandhini. "A study on optimization techniques in feature selection for medical image analysis." International Journal on Computer Science and Engineering (IJCSE) 9, no. 3 (2017).

[4] Deepa, S. N., and B. Aruna Devi. "A survey on artificial intelligence approaches for medical image classification." Indian Journal of Science and Technology 4, no. 11 (2011).

[5] Al-Wajih, Ranya, Said Jadid Abdulkadir, Norshakirah Aziz, Qasem Al-Tashi, and Noureen Talpur. "Hybrid binary grey wolf with Harris hawks optimizer for feature selection." IEEE Access 9 (2021).

[6] Salama, M. A., & Hassan, G. (2019). A Novel Feature Selection Measure Partnership-Gain. International Journal of Online and Biomedical Engineering (iJOE), 15(04), pp. 4–19. https://doi.org/10.3991/ijoe.v15i04.9831

[7] Abdel-Basset, Mohamed, Weiping Ding, and Doaa El-Shahat. "A hybrid Harris Hawks optimization algorithm with simulated annealing for feature selection." Artificial Intelligence Review54, no. 1 (2021): 593-637.

[8] K. Ren,W. Fang, J. Qu, X. Zhang, and X. Shi, ``Comparison of eight filter-based feature selection methods for monthly streamflow forecasting_Three case studies on CAMELS data sets,'' J. Hydrol., vol. 586, Jul. 2020,Art. no. 124897.

[9] M. T. Sadiq, X. Yu, Z. Yuan, F. Zeming, A. U. Rehman, I. Ullah, G. Li, and G. Xiao, ``Motor imagery EEG signals decoding by multivariate empirical wavelet transform-based framework for robust brain computer interfaces,'' IEEE Access, vol. 7, pp. 171431_171451, 2019.

[10] M. T. Sadiq, X. Yu, and Z. Yuan, ``Exploiting dimensionality reduction and neural network techniques for the development of expert brain_computer interfaces,'' Expert Syst. Appl., vol. 164, Feb. 2021.

[11] B. Xue, M. Zhang, W. N. Browne, and X. Yao, ``A survey on evolutionary computation approaches to feature selection,'' IEEE Trans. Evol. Comput.,vol. 20, no. 4, pp. 606_626, Aug. 2016.

[12] E.-G. Talbi, Metaheuristics: From Design to Implementation. Hoboken, NJ, USA: Wiley, 2009.

[13] Al-Wajih, Ranya, Said Jadid Abdulkadir, Norshakirah Aziz, Qasem Al-Tashi, and Noureen Talpur. "Hybrid binary grey wolf with Harris hawks optimizer for feature selection." IEEE Access 9 (2021): 31662-31677.

[14] Abdel-Basset, Mohamed, Weiping Ding, and Doaa El-Shahat. "A hybrid Harris Hawks optimization algorithm with simulated annealing for feature selection." Artificial Intelligence Review54, no. 1 (2021): 593-637.

[15] Thawkar, Shankar. "A hybrid model using teaching–learning-based optimization and Salp swarm algorithm for feature selection and classification in digital mammography." Journal of Ambient Intelligence and Humanized Computing 12, no. 9 (2021): 8793-8808.

[16] Ibrahim, Rehab Ali, Ahmed A. Ewees, Diego Oliva, Mohamed Abd Elaziz, and Songfeng Lu. "Improved salp swarm algorithm based on particle swarm optimization for feature selection." Journal of Ambient Intelligence and Humanized Computing 10, no. 8 (2019): 3155-3169.

[17] A. A. Heidari, S. Mirjalili, H. Faris, I. Aljarah, M. Mafarja, and H. Chen, ``Harris hawks optimization: Algorithm and applications,''Future Gener. Comput. Syst., vol. 97, pp. 849872, Aug. 2019

[18] Mirjalili, Seyedali, Amir H. Gandomi, Seyedeh Zahra Mirjalili, Shahrzad Saremi, Hossam Faris, and Seyed Mohammad Mirjalili. "Salp Swarm Algorithm: A bio-inspired optimizer for engineering design problems." Advances in Engineering Software 114 (2017): 163-191.

[19] Henschke, Natasha, Jason D. Everett, Anthony J. Richardson, and Iain M. Suthers. "Rethinking the role of salps in the ocean." Trends in Ecology & Evolution 31, no. 9 (2016): 720-733.

[20] Gupta, Shubham, Kusum Deep, Ali Asghar Heidari, Hossein Moayedi, and Mingjing Wang. "Opposition-based learning Harris hawks optimization with advanced transition rules: Principles and analysis." Expert Systems with Applications158 (2020): 113510.

[21] Radiological Society of North America (RSNA), "Radiological Society of North America (RSNA)," Radiological Society of North America (RSNA), 2020. https://www.rsna.org/

[22] R. Dataset, "Radiopaedia dataset," 2020. https://radiopaedia.org/articles/imaging-data-sets-artificial-intelligence

[23] C. G. Repository, "Cohen's GitHub repository," *Cohen's GitHub repository*, 2021. https://github.com/ieee8023

[24] Italian Society of Medical and Interventional Radiology (SIRM), "Italian Society of Medical and Interventional Radiology (SIRM)," Italian Society of Medical and Interventional Radiology (SIRM), 2020. https://sirm.org/la-radiologia-medica/

[25] Kaggle, "Kaggle's chest X-ray images (Pneumonia) dataset," Kaggle's chest X-ray images (Pneumonia) dataset, 2020. https://www.kaggle.com/datasets/paultimothymooney/chest-xray-pneumonia

[26] Abdel-Basset, Mohamed, Weiping Ding, and Doaa El-Shahat. "A hybrid Harris Hawks optimization algorithm with simulated annealing for feature selection." Artificial Intelligence Review 54, no. 1 (2021): 593-637.

[27] Fusilier, Donato Hernández, Manuel Montes-y-Gómez, Paolo Rosso, and Rafael Guzmán Cabrera. "Detecting positive and negative deceptive opinions using PU-learning." Information processing & management 51, no. 4 (2015): 433-443.

[28] Chen, Tianqi, and Carlos Guestrin. "Xgboost: A scalable tree boosting system." In Proceedings of the 22nd acm sigkdd international conference on knowledge discovery and data mining, pp. 785-794. 2016.

[29] Yasir, M. A., & Ali, Y. H. (2021). Review on Real Time Background Extraction: Models, Applications, Environments, Challenges, and Evaluation Approaches. International Journal of Online and Biomedical Engineering (iJOE), 17(02), pp. 37–68. https://doi.org/10.3991/ijoe.v17i02.18013

## 11   Authors

**Abubakr S. Issa** received his bachelor's degree in computer science department – Artificial intelligence branch from the University of Technology (UOT) – Iraq 2014. Since 2014, he is working as a programmer at the Information Technology Center, at the University of Technology up till now. Meanwhile, he is an M.Sc candidate at the University of Technology (UOT) – Iraq.

**Assistant Professor Dr. Yossra Hussain Ali**. She received her B.Sc, M.Sc, and Ph.D. degrees in 1996, 2002, and 2006 respectively from Iraq, the University of Technology, Department of Computer Sciences. She joined the University of Technology, Iraq in 1997. During her postgraduate studies, she worked on Computer Networks, Information systems, Agent Programming and Image Processing as well as some experience in Artificial Intelligent and Computer Data Security. She is a reviewer at many conferences and journals and she supervised several undergraduate and postgraduate (PhD. and MSc.) dissertations in Computer sciences. Yossra has many professional certificates and she has published in well-regarded journals (e-mail: yossra.h.ali@uotechnology.edu.iq).

**Tarik A. Rashid** received his Ph.D. in Computer Science and Informatics degree from the College of Engineering, Mathematical and Physical Sciences, University College Dublin (UCD) in 2001–2006. He pursued his Post-Doctoral Fellow at the Computer Science and Informatics School, College of Engineering, Mathematical and Physical Sciences, University College Dublin (UCD) from 2006–2007. He joined the University

Issa, A., Ali, Y., & Rashid , T. (2022). An Efficient Hybrid Classification Approach for COVID-19 Based on Harris Hawks Optimization and Salp Swarm Optimization. *International Journal of Online and Biomedical Engineering (iJOE)*, *18*(13), pp. 113–130. https://doi.org/10.3991/ijoe.v18i13.33195

of Kurdistan Hewlêr (UKH) in 2017. He has also been included in the prestigious Stanford University list of 2.% of the best world researchers for the years 2020 and 2022.